\documentclass[10pt,twocolumn,letterpaper]{article}

\usepackage{cvpr}
\usepackage{times}
\usepackage{epsfig}
\usepackage{graphicx}
\usepackage{amsmath}
\usepackage{amssymb}
\usepackage{color}
\usepackage{multirow}

\usepackage[breaklinks=true,bookmarks=false]{hyperref}

\cvprfinalcopy 

\begin{document}

\title{3DSRnet: Video Super-resolution using 3D Convolutional Neural Networks}

\author{Soo Ye Kim\\
KAIST\\
{\tt\small sooyekim@kaist.ac.kr}
\and
Jeongyeon Lim\\
SK Telecom\\
{\tt\small jeongyeon@sk.com}
\and
Taeyoung Na\\
SK Telecom\\
{\tt\small taeyoung.na@sk.com}
\and
Munchurl Kim\\
KAIST\\
{\tt\small mkimee@kaist.ac.kr}
}

\maketitle

\begin{abstract}
In video super-resolution, the spatio-temporal coherence between, and among the frames must be exploited appropriately for accurate prediction of the high resolution frames. Although 2D convolutional neural networks (CNNs) are powerful in modelling images, 3D-CNNs are more suitable for spatio-temporal feature extraction as they can preserve temporal information. To this end, we propose an effective 3D-CNN for video super-resolution, called the 3DSRnet that does not require motion alignment as preprocessing. Our 3DSRnet maintains the temporal depth of spatio-temporal feature maps to maximally capture the temporally nonlinear characteristics between low and high resolution frames, and adopts residual learning in conjunction with the sub-pixel outputs. It outperforms the most state-of-the-art method with average 0.45 and 0.36 dB higher in PSNR for scales 3 and 4, respectively, in the Vidset4 benchmark. Our 3DSRnet first deals with the performance drop due to scene change, which is important in practice but has not been previously considered. 

\end{abstract}

\section{Introduction}

Vision is one of the most primitive yet sophisticated sensory systems that is continuously stimulated not only by natural scenes, but also by electric displays. With the unceasingly evolving display hardware which has now commercially reached the resolution of 8K Ultra High Definition (UHD), and people's rising expectations on these types of visuals, the demand for better quality videos is at its highest. However, the sole advancement of display technologies is not sufficient to offer high quality visual content to users - the contents themselves have to be of higher resolution. Although they can be obtained through the usage of high-end filming equipment, it is costly and problematic due to large storage and transmission bandwidth required.

Super-resolution (SR) is an imaging technique that transforms low resolution (LR) images to higher resolution ones. When an LR image is given as input, an SR algorithm exploits its internal information to generate an output image, hopefully similar to its high resolution (HR) counterpart. This is regarded as an ill-posed problem since multiple HR images correspond to a single LR image. Non-existent, but reasonable, information should be created within the image when going from LR to HR, and finding a high quality image among the possible solutions is the key to the SR problem.

\begin{figure}
\centering
\includegraphics[scale=0.45]{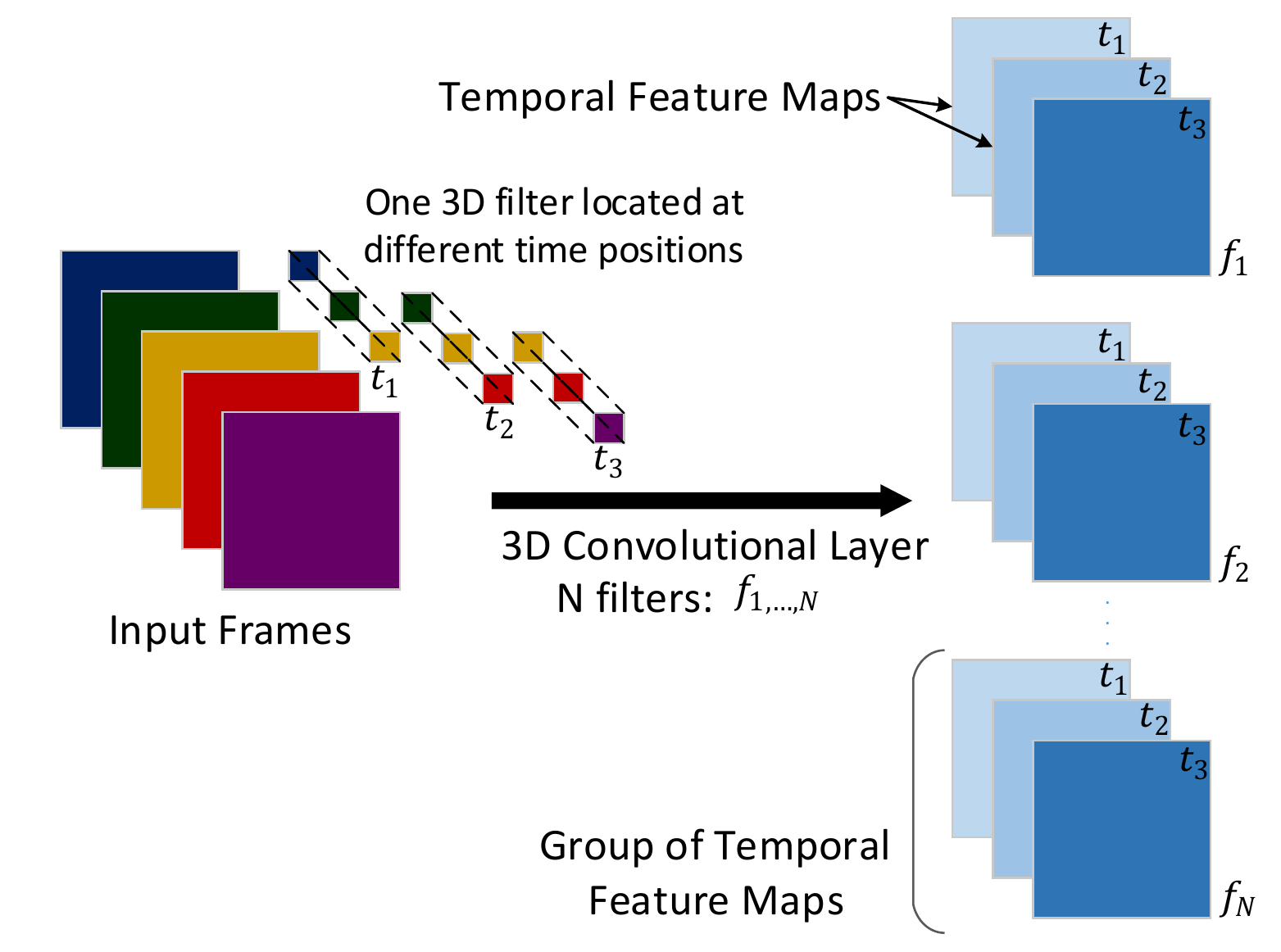}
\caption{Illustration of a 3D convolution layer with a five-frame input and filters of depth 3. When \textit{N} 3D filters are applied to the input, each of the filters generate a group of temporal feature maps, resulting in \textit{N} groups of temporal feature maps}
\label{fig:1}
\end{figure}

Despite that SR is a popular problem in image processing and computer vision, most studies have focused on single image SR than multi-frame SR, also referred to as video SR. However, many SR applications are in videos where the reconstruction of HR frames may benefit from additional information contained in the previous and future LR frames. While video frames exhibit high temporal coherence, the camera or object motion can also provide a different angle or scale of the parts in the current frame in the consecutive surrounding frames, which can be effectively utilized as crucial clues in constructing high quality HR frames.

A video SR algorithm should fully exploit the temporal relations between the consecutive frames to aggregate them with the spatial information. To this end, we propose an effective 3D convolutional neural network (CNN) for video SR, called 3DSRnet that does not require motion estimation nor compensation to interpret the spatio-temporal information in consecutive frames. Instead, it finds an end-to-end nonlinear spatio-temporal mapping in itself through residual learning and lowers complexity by using the multi-channel output structure introduced in \cite{shi2016real}. Our 3DSRnet outperforms the previous video SR methods \cite{caballero2017real,kappeler2016video} by at least average 0.36 dB in PSNR for the Vidset4 benchmark test dataset. To the best of our knowledge, it is also the first video SR method that can effectively deal with scene change in the input frames.

\section{Related Work}

\subsection{Single Image Super-resolution}

Single image SR attempts to develop an HR image from a single LR image. Past attempts to tackle this problem include internal and external example-based methods \cite{timofte2013anchored,timofte2014a+,glasner2009super,yang2010image,schulter2015fast}. The former includes a method devised by Glasner et al. \cite{glasner2009super}, which identifies internal redundancies in an image to obtain essential information in upscaling of the patches. The external example-based methods try to find a dictionary mapping \cite{timofte2013anchored,timofte2014a+,schulter2015fast}. Another type of approach is through sparse representation, applied successfully by Yang et al \cite{yang2010image}. 

With the recent rise of deep learning and the excellent performance of CNNs in image classification \cite{krizhevsky2012imagenet}, the first structure that adopts a CNN structure for SR was proposed by Dong et al. \cite{dong2016image,dong2014learning}, which suggests a simple 3-layer structure. Their model, called SRCNN, demonstrated great potential of using CNNs for SR applications. Since then, CNN-based structures have been boasting superior performance. One of the CNN-based SR methods that was highly successful is called the very deep super-resolution method (VDSR) proposed by Kim et al \cite{kim2016accurate}. The VDSR has as many as twenty convolution layers and first adopts residual learning to train a deep SR network. However, both SRCNN and VDSR start with enlarged LR images using a bicubic filter, as input to the first convolution layer. Consequently, the convolution operations are taken place on the enlarged input, which leads to high computation complexity. An inspirational work by Shi et al. \cite{shi2016real} suggested a sub-pixel CNN that finds a direct transform from the LR image by using the fact that convolution layers can produce multiple channels at the output. With this multi-channel output structure, the HR image can be obtained through a simple reordering of the output pixels. Our 3DSRnet employs this multi-channel output structure \cite{shi2016real} with residual learning in \cite{kim2016accurate}.

\subsection{Video Super-resolution}

Video SR, or multi-frame SR, assumes that the input is a series of consecutive frames at each time instance of video sequences. Undoubtedly, single image SR algorithms may be applied on the individual frames for videos, and this may even be more efficient in some cases if they achieve real-time performance as in \cite{shi2016real}. However, more spatial information is available in the case of videos, as not only the current LR frame but also its surrounding consecutive LR frames may be utilized. This means that to fully profit from what is given, the temporal relation of the spatial information has to be carefully taken into account in reconstructing the corresponding HR frame.

Compared to image SR, relatively less studies have been conducted on video SR. Focusing on neural network based methods, Kappeler et al. \cite{kappeler2016video} extended SRCNN to 2D-CNN architectures that combine information from neighboring frames. Caballero et al. \cite{caballero2017real} proposed three video SR architectures where the early and slow fusion architectures have a similar way of dealing with the multi-frame input as in \cite{kappeler2016video,karpathy2014large}. The structures in \cite{caballero2017real} all adopt the same multi-channel output structure in \cite{shi2016real}. The third model is a 3D-CNN architecture that first incorporated 3D convolution filters into video SR to capture temporal information of multiple frames. This model was a conceptual suggestion without the specific configuration information presented, where the temporal depth of feature maps shrinks to one in early convolution layers on which 2D convolutions are then performed. No performance comparison for the 3D-CNN architecture \cite{caballero2017real} was provided against other previous methods due to its relatively lower SR performance compared to the early and slow fusion architectures. 

In comparison, our 3DSRnet maintains the temporal depth of spatio-temporal feature maps towards deeper layers to maximally capture the temporally nonlinear characteristics between LR and HR frames, and we provide intensive experiments and analysis on 3DSRnet in the later sections of this paper. All three video SR architectures in \cite{caballero2017real} and the video SR method in \cite{kappeler2016video} need motion alignment among the multiple input frames whereas our 3DSRnet directly takes the input frames without any motion alignment. None of the previous CNN-based video SR methods has considered scene change issues while our 3DSRnet first incorporates a scene change detection network to locate a scene change boundary in multiple input frames and replace the different scene frames with the temporally closest frame of the same scene as the current frame scene.

\begin{figure*}
\centering
\includegraphics[width=1\textwidth]{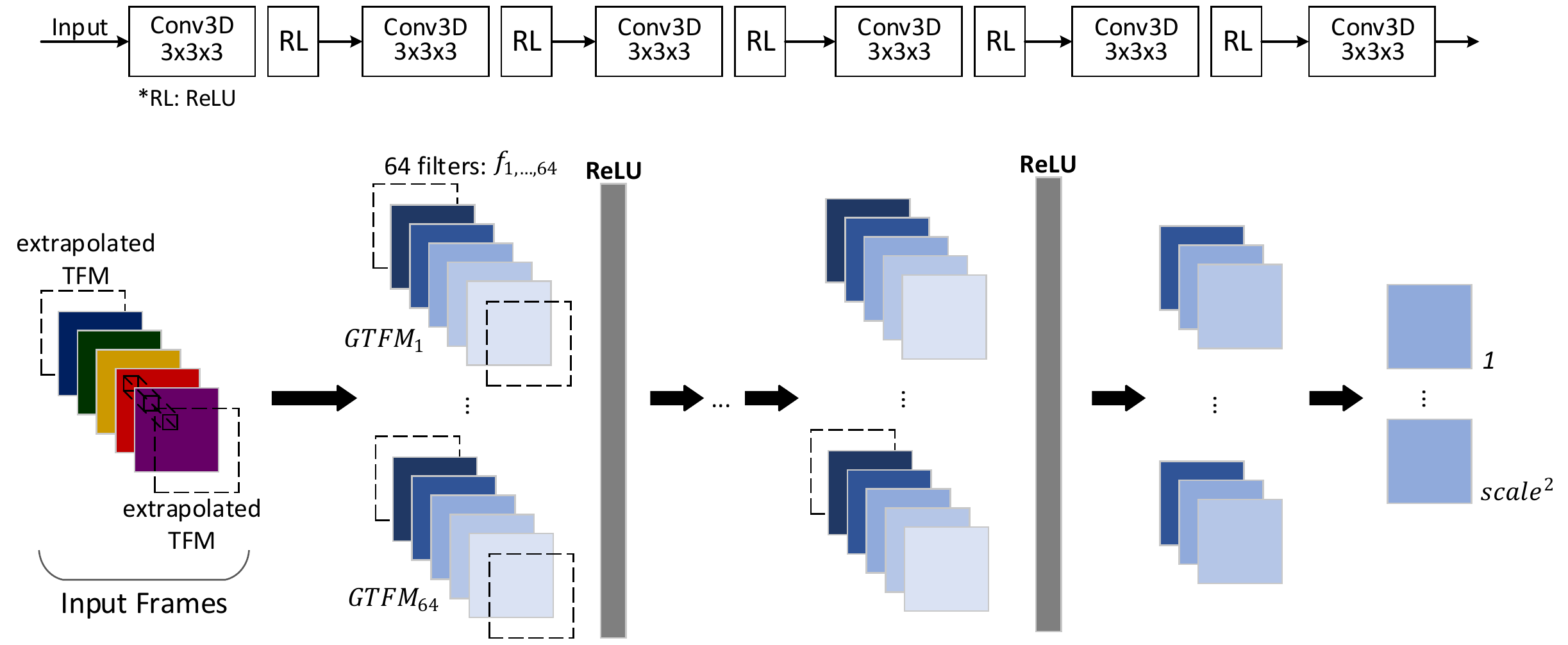}
\caption{Architecture of the video SR subnet in 3DSRnet. Each convolution layer has 64 filters of size 3$\times$3$\times$3 and is followed by ReLU activation except the last layer. Each filter produces a group of temporal feature maps (GTFM) and a temporal feature map is extrapolated on both ends of each GTFM to preserve the temporal depth which would otherwise decrease. No extrapolation is performed from layer \textit{L-1} so that the temporal information is merged to produce the final 2D output. The temporal information is preserved until the last layer}
\label{fig:2}
\end{figure*}

\subsection{3D Convolutional Neural Networks}

2D-CNN, a common neural network used for images, is a powerful structure in modelling images with their spatial feature extraction capability. However when another axis, time, is introduced as in videos, we argue that 3D-CNN is a more suitable option for spatio-temporal feature extraction. This is in line with Tran et al. \cite{tran2015learning} in which they argued that 3D-CNN is an effective video descriptor and with \cite{ji20133d} that demonstrated the spatial and temporal feature extraction capability of 3D-CNNs. 3D-CNNs have been successfully implemented in high level vision tasks for videos such as action/object recognition and scene/event classification \cite{tran2015learning,ji20133d,teivas2016video}. We believe they are also effectively applicable to a low level vision task for videos such as video SR. In this paper, we adopt the 3D-CNN and design an elaborate video SR network, 3DSRnet, which makes the 3D convolution effective on video SR where motion alignment is not necessitated thanks to its spatio-temporal feature representation ability.

\section{Proposed Method}

We propose a 3D-CNN architecture for video SR named as the 3DSRnet with an additional scene change module that deals with scene change occurring inputs. The 3DSRnet consists of two subnets: \smallskip \\ \smallskip \indent (i) Video SR subnet, and \\ \indent (ii) Scene change detection and frame replacement (SF) subnet. \smallskip \\The video SR subnet takes a series of consecutive LR input frames in a sliding time window, and produces an HR output frame corresponding to the middle frame in the sliding time window. The SF subnet of the 3DSRnet is responsible for the detection of scene change in the sliding time window, and replaces the frames of a different scene with the temporally closest frame that belongs to the same scene as the middle frame.

\subsection{Video Super-resolution Subnet}

\subsubsection{3D Convolution Layers.} 

The video SR subnet is composed of 3D convolution layers where 3D filters of size $height \times width \times depth$ are applied on the input composed of multiple consecutive frames or feature maps. Unlike 2D filters of size $height \times width$ that are applied on the full depth of the input and slid horizontally and vertically, 3D filters have a third size parameter, \textit{depth}, so that they are swept horizontally, vertically and depth-wise. The first 3D convolution layer takes a series of five consecutive frames in a sliding input window where each 3D filter generates a temporal feature map (TFM) at the corresponding frame position, each filter yielding a group of temporal feature maps (GTFMs) from all the input frame positions. This is illustrated in Fig. \ref{fig:1} where \textit{N} 3D filters, $f_{1,...,N}$ of depth 3 are applied on the input composed of five frames to produce \textit{N} GTFMs. The temporal depth of each GTFM is 3. Formally, the \textit{n}-th GTFM before activation in the first 3D convolution layer is given by 
\begin{align}  
	GTFM_{xyt}^{n} & = \sum_{h=1}^{H}\sum_{w=1}^{W}\sum_{d=1}^{D} w_{hwd}^{n}v_{\left(x+h\right)\left(y+w\right)\left(t+d\right)} + b^n
\end{align}
where $w_{HWD}^n$is the 3D filter \textit{n} of size $H \times W \times D$ and \textit{v} is the input. 

\begin{figure*}
\centering
\includegraphics[width=1\textwidth]{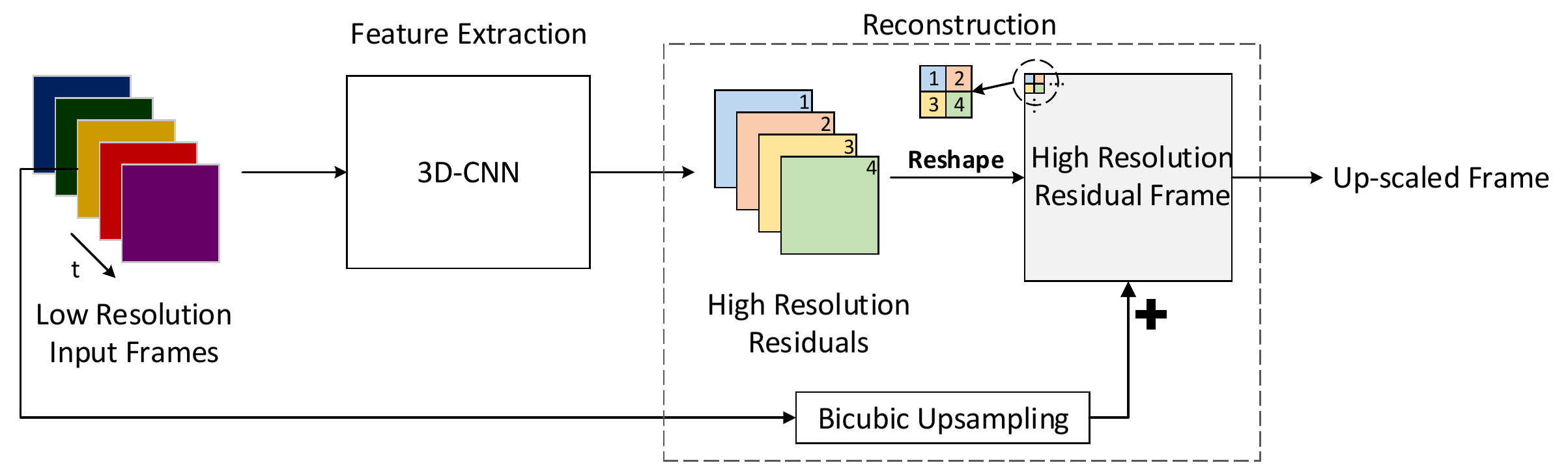}
\caption{The input and output structure of 3DSRnet. This is an example with five input frames and the scale factor of 2. The input frames go through a 3D-CNN for spatio-temporal feature extraction and this 3D-CNN predicts the high resolution residuals of the middle frame. There are four output channels because the scale factor is 2 (\textit{no. of output channels=$scale^2$}). They are reshaped to make the high resolution residual frame. The final up-scaled frame is obtained by adding the bicubic up-scaled middle frame to the residual frame}
\label{fig:3}
\end{figure*}

From the second to the last 3D convolution layer, the input window corresponds to the whole set of multiple GTFMs where each GTFM is generated from one 3D filter of the previous convolution layer. The temporal information contained within the input window is preserved through the 3D convolution layer in each GTFM as separate TFMs unlike 2D convolution layers where the input would be collapsed into one single feature map per filter. From the second to the last 3D convolution layer, the input \textit{v} is composed of multiple GTFMs (\textit{m} GTFMs), and the \textit{n}-th GTFM before activation is given by
\begin{align}  
	GTFM_{xyt}^{n} & = \sum_{m}\sum_{h=1}^{H}\sum_{w=1}^{W}\sum_{d=1}^{D} w_{hwd}^{mn}v_{\left(x+h\right)\left(y+w\right)\left(t+d\right)}^{m} + b^n.
\end{align} 

We use the ReLU \cite{glorot2011deep} function as the activation function after every convolution layer except the last, for the nonlinearity of the network. In 3D-CNNs, temporal nonlinearities among the GTFMs as well as spatial nonlinearities are introduced thanks to the 3D convolution structure.

\subsubsection{Extrapolation.} 
The temporal depth of GTFMs become shallower as the network gets deeper, as the 3D filters integrate the temporal information. For example, with an input of five frames and a 3D filter of depth 3, the output GTFMs would have depth 1 after only two 3D convolution layers, being no different from a 2D convolution layer from layers thereafter. Since the usage of 3D-CNNs is to tamper with the temporal information, thereby introducing temporal nonlinearities, extrapolating (or padding) the input GTFMs at their front and back ends allows to preserve the temporal depth throughout the network. However, for the last layers, no extrapolation is performed and the temporal information is aggregated, to produce the final 2D HR frame as intended. For an input of five frames and a 3D filter of depth 3, no extrapolation is carried out from layer \textit{L-1} where \textit{L} is the number of convolution layers. This naturally aggregates the temporal information when going deeper in the network. Please refer to Fig. \ref{fig:2} for a detailed illustration of the 3D convolution layers with extrapolation.

\subsubsection{Multi-channel Output.} 
The multi-channel output structure first introduced in \cite{shi2016real} allows for a direct mapping from the LR to HR frames by producing an output with multiple channels that can simply be reordered and reshaped to produce the final HR output. This method alleviates the amount of computation which can be otherwise expensive for 3D-CNNs. Furthermore, it can enhance SR performance because the receptive field of the LR input pixels without bicubic up-scaling is larger than that of an up-scaled LR input pixels, provided that the filter size and network depth are the same. Large receptive fields are essential in SR to yield high performance \cite{kim2016accurate,huang2017image,wang2017large,shi2017single}.

\subsubsection{Residual Learning.} 
HR frames often consist of low and high frequency components. However, the low frequency components are mostly present in the LR frames, meaning that the essential goal of an SR algorithm is in predicting the missing high frequency components. Therefore, the network can save the trouble of predicting what is already there, by directly predicting the difference between the HR frame and the corresponding bicubic-upscaled LR frame - the \textit{residual} frame -. Our 3DSRnet employs this technique and predicts the residual frame, producing a \textit{multi-channel residual} output.  Residual learning was first proposed in \cite{he2016deep} and applied to SR in \cite{kim2016accurate}. It also eases training \cite{kim2016accurate} by solving the vanishing and exploding gradient problem which can be critical in training neural networks \cite{bengio1994learning}. Fig. \ref{fig:3} shows the input and output structures of the 3DSRnet.

\begin{figure*}
\centering
\includegraphics[width=1\textwidth]{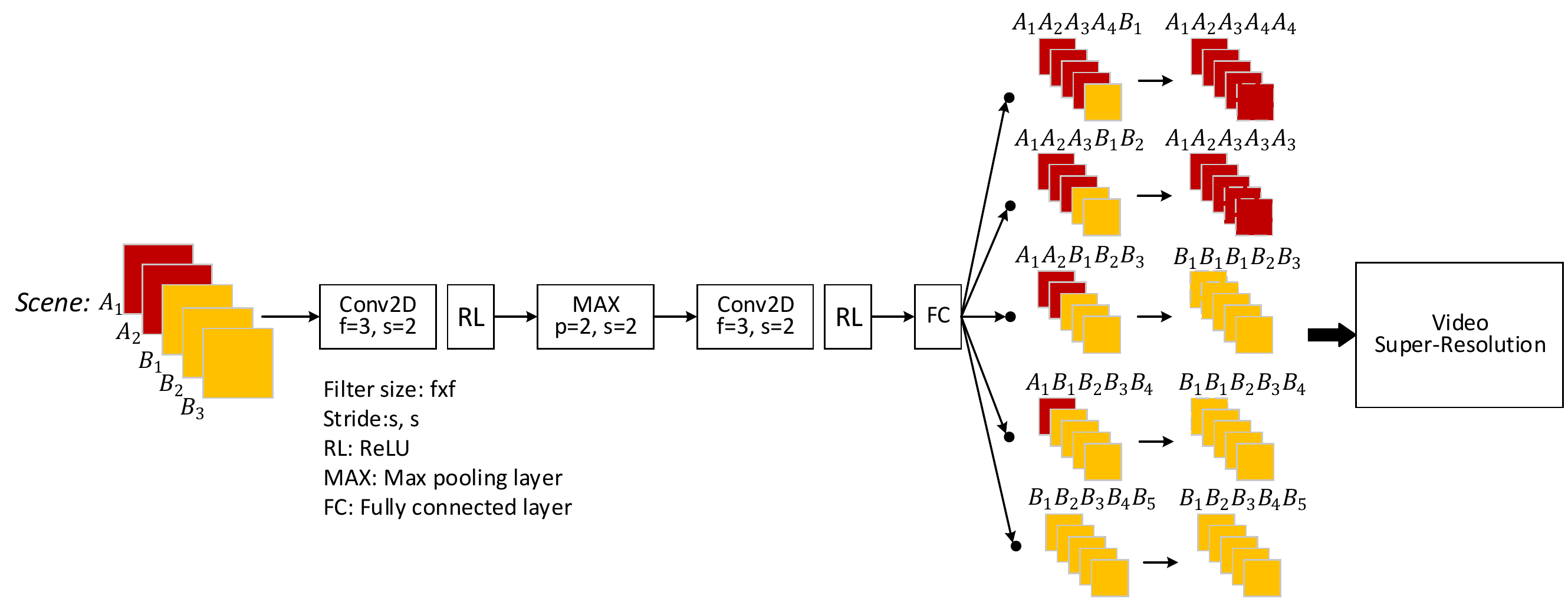}
\caption{Illustration of the 3-layer SF subnet for a sliding time window with five consecutive input frames. The squares in red and yellow colors denote the frames of scene A and scene B, respectively. The input is classified into whether a scene change occurs after frame index 1, 2, 3, 4 or not. After classification, the frames of a different scene are replaced with the temporally closest frame that belongs to the same scene as the middle frame. Note that the middle frame is considered as the reference, and the frames of a different scene than the middle frame are swapped with that of the same scene}
\label{fig:4}
\end{figure*}

\subsection{Scene Change Detection and Frame Replacement Subnet}

The scene change detection and frame replacement (SF) subnet is another component of the 3DSRnet. When multiple frames are used as input to video SR networks, there is a possibility of scene change within them. In this case, the performance of a video SR algorithm drops due to the frames of different scenes getting involved into convolution, resulting in the reconstructed HR frames of poor quality. The previous video SR methods avoided this problem by explicitly collecting data without scene changes, which is impractical in real world applications. Our 3DSRnet handles the scene change problem for video SR by introducing the SF subnet that classifies the exact location of the scene boundary and modifies some of the frames in the sliding input window by replacing the different scene frames with the temporally closest frame of the same scene as the current frame scene. Although the duplicated (replaced) frames do not contain any new information, this method significantly helps the 3DSRnet alleviate performance degradation from the contaminated input of different scene frames.

\subsubsection{Identification of Scene Change Location.} 
If we assume that a scene change may occur within the five input frames, there are four possible scene change locations (labels). In addition, the fifth label is designated for no scene change. Then it is a simple five-class classification problem. Fig. \ref{fig:4} illustrates the detailed mechanism of the SF subnet for a sliding time window of five consecutive input frames. The SF subnet should be lightweight as it can be optionally used alongside video SR, but accurate to correctly modify the input. Therefore, we use a shallow 2D-CNN structure. It is trained separately from the video SR subnet.

\section{Experiments}
\subsection{Experiment Conditions}

\begin{table} [b]
\begin{center}
\caption{Training data sets for the video SR subnet}
\label{table:1}
\scalebox{0.86}{
\begin{tabular}{ c|c|c|c|c|@{}c@{} }
\hline \hline
\multirow{2}{*}{Dataset} & \multicolumn{2}{c|}{Type 1} & \multicolumn{2}{c|}{Type 2} & {Total no.} \\
\cline{2-5}
& stride & subim./fr. & stride & subim./fr. & of subim.\\
\hline \hline
\textit{smallSet}&30&5&5&10&9,725\\
\hline
\textit{largeSet}&15&10&5&12&14418\\
\hline \hline
\multicolumn{6}{l}{*subimages and frame are shortened as subim and fr, respectively.}
\end{tabular}}
\end{center}
\end{table}

\begin{table*}
\begin{center}
\caption{Experiment on architectures trained on the \textit{smallSet}}
\label{table:2}
\scalebox{0.9}{
\begin{tabular}{ c|c|c|c|c|c|c|c|c|c|c }
\hline \hline
&\multicolumn{2}{c|}{2D-CNN} & \multicolumn{2}{c|}{3DSRnet v1} & \multicolumn{2}{c|}{3DSRnet v2} & \multicolumn{2}{c|}{3DSRnet v3} & \multicolumn{2}{c}{3DSRnet}\\
\hline
Layers & \multicolumn{10}{c}{Number of filter channels (input, output)} \\
\hline
1&2D&5, 32&3D&1, 32&3D&1, 32&3D&1, 32&3D&1, 32 \\
2&2D&32, 64&3D&32, 32&3D&32, 32&3D&32, 32&3D&32, 32 \\
3&2D&64, 64&3D&32, 16&3D&32, 32&3D&32, 32&3D&32, 32 \\
4&2D&64, 64&2D&80, 64&3D&32, 16&3D&32, 32&3D&32, 32 \\
5&2D&64, 35&2D&64, 32&2D&80, 64&3D&32, 16&3D&32, 32 \\
6&2D&35, 4&2D&32, 4&2D&64, 4&2D&80, 4&2D&32, 4 \\
\hline
2D filter size & \multicolumn{2}{c|}{3$\times$3} & \multicolumn{2}{c|}{3$\times$3} & \multicolumn{2}{c|}{3$\times$3} & \multicolumn{2}{c|}{3$\times$3} & \multicolumn{2}{c}{3$\times$3}\\
3D filter size & \multicolumn{2}{c|}{-} & \multicolumn{2}{c|}{3$\times$3$\times$3} & \multicolumn{2}{c|}{3$\times$3$\times$3} & \multicolumn{2}{c|}{3$\times$3$\times$3} & \multicolumn{2}{c}{3$\times$3$\times$3}\\
\hline
Concat. layer&\multicolumn{2}{c|}{-}&\multicolumn{2}{c|}{3}&\multicolumn{2}{c|}{4}&\multicolumn{2}{c|}{5}&\multicolumn{2}{c}{-}\\
\hline \hline
Total parameters&\multicolumn{2}{c|}{115,020}&\multicolumn{2}{c|}{108,000}&\multicolumn{2}{c|}{118,368}&\multicolumn{2}{c|}{100,512}&\multicolumn{2}{c}{114,912}\\
\hline
PSNR (dB)&\multicolumn{2}{c|}{32.49}&\multicolumn{2}{c|}{32.81}&\multicolumn{2}{c|}{32.89}&\multicolumn{2}{c|}{32.85}&\multicolumn{2}{c}{\textbf{32.92}}\\

\hline \hline
\end{tabular}}
\end{center}
\end{table*}

\begin{table*}
\begin{center}
\caption{Vidset4 Benchmark-Video SR Methods}
\label{table:3}
\scalebox{0.9}{
\begin{tabular} {c|c|c|c|c|c|c}
\hline \hline
Vidset4&Bayesian \cite{liu2014bayesian}&Deep-DE \cite{liao2015video}&VSRnet \cite{kappeler2016video}&Liu \textit{et al.} \cite{Liu2017}&VESPCN \cite{caballero2017real}& 3DSRnet\\
\hline \hline
$\times$4&24.66&24.68&24.84&25.24&25.35&\textbf{25.71}\\
\hline \hline
\end{tabular}}
\end{center}
\end{table*}

\subsubsection{Data.} A training or testing data sample of 3DSRnet is composed of five bicubic-down-scaled LR frames and a single HR middle frame. We collected two sets of 3840$\times$2160 UHD videos of 30 fps that were encoded with at least 100 Mb/s using an H.264/AVC encoder. The first video set (Type 1) shows spatially complex scenes, meaning that they contain sophisticated objects such as the bird view of a city, and the second video set (Type 2) is temporally complex, meaning that there is a lot of motion. We collected three Type 1 videos of total 8,504 frames and one Type 2 video of 8,655 frames. They were converted into 420 YUV format and only the Y channel was used as the training and test data. When reconstructing color frames, U and V channels were simply up-scaled using a bicubic filter.

For training the video SR subnet in 3DSRnet, we prepared two datasets, \textit{smallSet} and \textit{largeSet}, where a predefined number of non-overlapping subimages were randomly selected from each frame with a frame stride from Type 1 and Type 2 sets. For fair comparison with other video SR methods, the video SR subnet was trained with a training dataset without scene change. Table \ref{table:1} summarizes the training sets for the video SR subnet of the 3DSRnet. The size of LR subimages for the scale factors 2, 3 and 4 were 80, 60, 40 for the \textit{smallSet} and 80, 60, 45 for the \textit{largeSet}, respectively. The training took around three days with the \textit{smallSet} and eight days with the \textit{largeSet} using an Nvidia TITAN X GPU for a scale factor of 2. For the comparison among the video SR subnet of 3DSRnet and its variants, the test set contains the data samples of scenes that are not included in the training set. To compare the video SR subnet of 3DSRnet with the state-of-the-art SR methods, we used the Vidset4 dataset which is a commonly used test set for videos. 

For training the SF subnet in 3DSRnet, a separate dataset was created to contain scene changes. The LR frames of different scenes from the smallSet were reduced by a factor of 40 to be of size 48$\times$27, and randomly concatenated to make scene change occurring inputs. 2,000 data samples each consisting of the frames and its label were randomly selected for each of the five classes to make the final training data of 10,000 data samples.

\subsubsection{Training.} 
The purpose of the video SR subnet was to minimize the mean squared loss between the predicted frame $F\left(X_i; \theta\right)$ and the ground truth frame $Y_i$, given by
\begin{align}  
	Loss\left(\theta\right) = \frac{1}{2n}\sum_{i=1}^{n}{\lVert}F\left(X_i; \theta\right)-Y_i{\rVert}^2
\end{align}
where $X_i$ is the input frames, $\theta$ is the set of model parameters and \textit{n} is the number of data samples. Then the gradient is calculated as the difference between $F\left(X_i; \theta\right)$ and $Y_i$. All weights were initialized by the Xavier initialization \cite{glorot2010understanding} using both the number of input and output neurons of the layer. The parameters were updated using Adam \cite{kingma2014adam}.

All 3D filter sizes were empirically set to 3$\times$3$\times$3, 2D filters to size 3$\times$3 and the number of filters are 64 if not otherwise mentioned. The network is composed of six convolution layers, considering the tradeoff between performance and complexity. The learning rate was set to 5$\times 10^{-4}$ for the \textit{smallSet} and $10^{-4}$ for the \textit{largeSet}. The learning rates of biases are 10 times smaller. For all network models, the weight decay was set to 5$\times 10^{-4}$ for filters and zero for biases. The mini-batch size is 32 for the \textit{smallSet} and 64 for the \textit{largeSet}. All models were implemented using the MatConvNet \cite{vedaldi2015matconvnet} package and 3D convolution layers were added using a Matlab mex implementation available in GitHub\footnote{https://github.com/pengsun/MexConv3D}.

\begin{figure}
\centering
\includegraphics[scale=0.6]{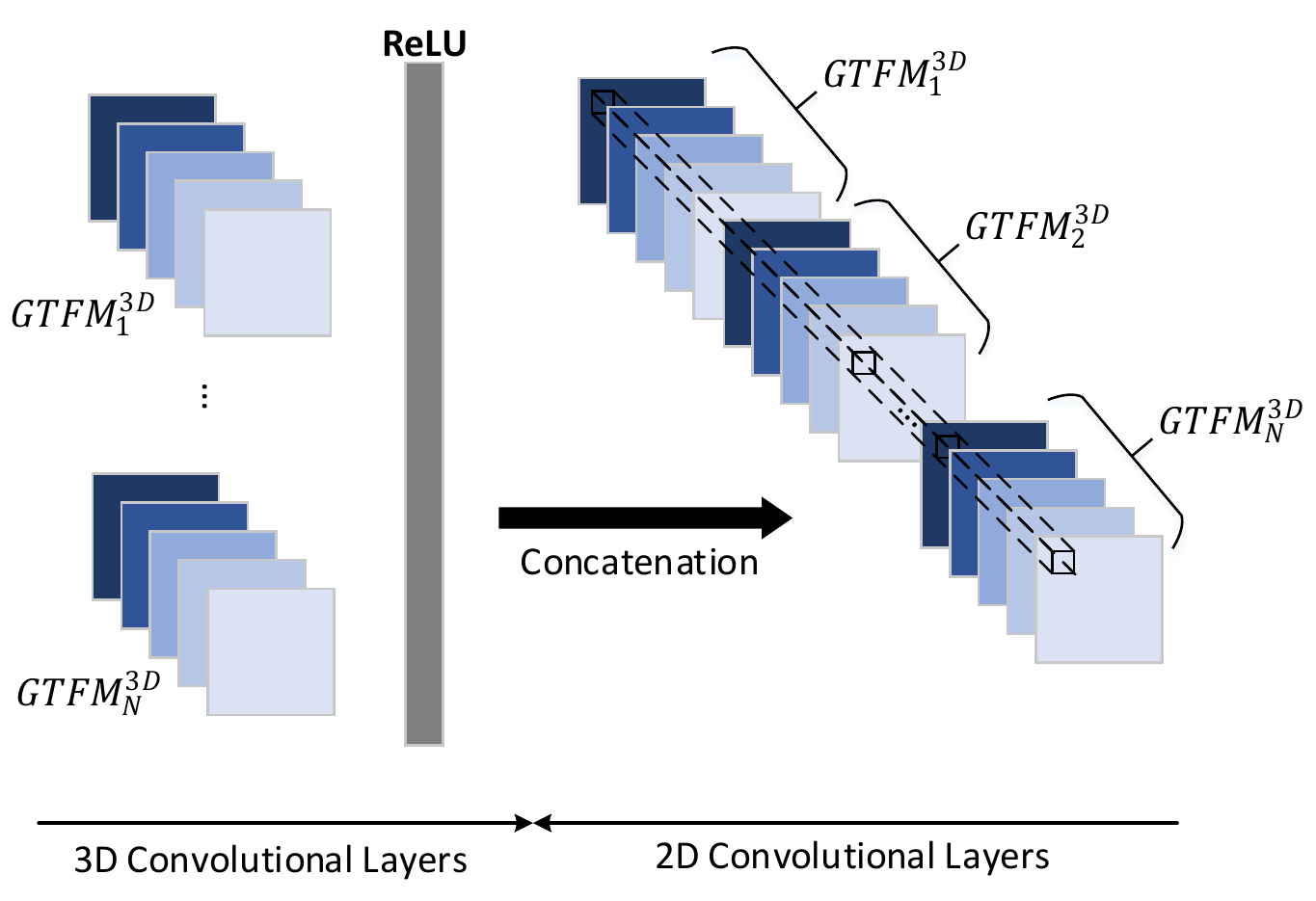}
\caption{Concatenation layer for 3D-2D integrated architectures. The GTFMs generated from the last 3D convolution layer are concatenated as input for the following first 2D convolution layer}
\label{fig:5}
\end{figure}

\subsection{Architecture}

\subsubsection{Effect of 3D-CNNs over 2D-CNNs.} The video SR subnet in 3DSRnet takes a 3D input (multiple LR input frames) in a sliding time window at a time instance and produces one single 2D HR output frame. So its architecture must be devised to go from 3D to 2D. As illustrated in in Fig. \ref{fig:2}, the temporal depth of the GTFMs is kept constant until the (\textit{L-2})-th convolution layer, and from the (\textit{L-1})-th convolution layer, no more temporal extrapolation is done to gradually reduce the temporal depth of GTFMs to 1 for our 3DSRnet. As variants of the 3DSRnet, we also experimented with the combination of 3D and 2D convolution layers by simply concatenating the GTFMs created from the last 3D convolution layer and performing 2D filtering thenceforth, with the number of filters adjusted so that all architectures have a similar number of parameters. The concatenation layer is illustrated in Fig. \ref{fig:5}. Table \ref{table:2} summarizes the specifications and results of our 3DSRnet and its variants, 3DSRnet v1, 3DSRnet v2 and 3DSRnet v3 with the comparison to a 2D-CNN structure, also with a five-frame input, made available to demonstrate the superior feature extraction capability of 3D-CNNs.

\begin{figure}
\centering
\includegraphics[scale=0.45]{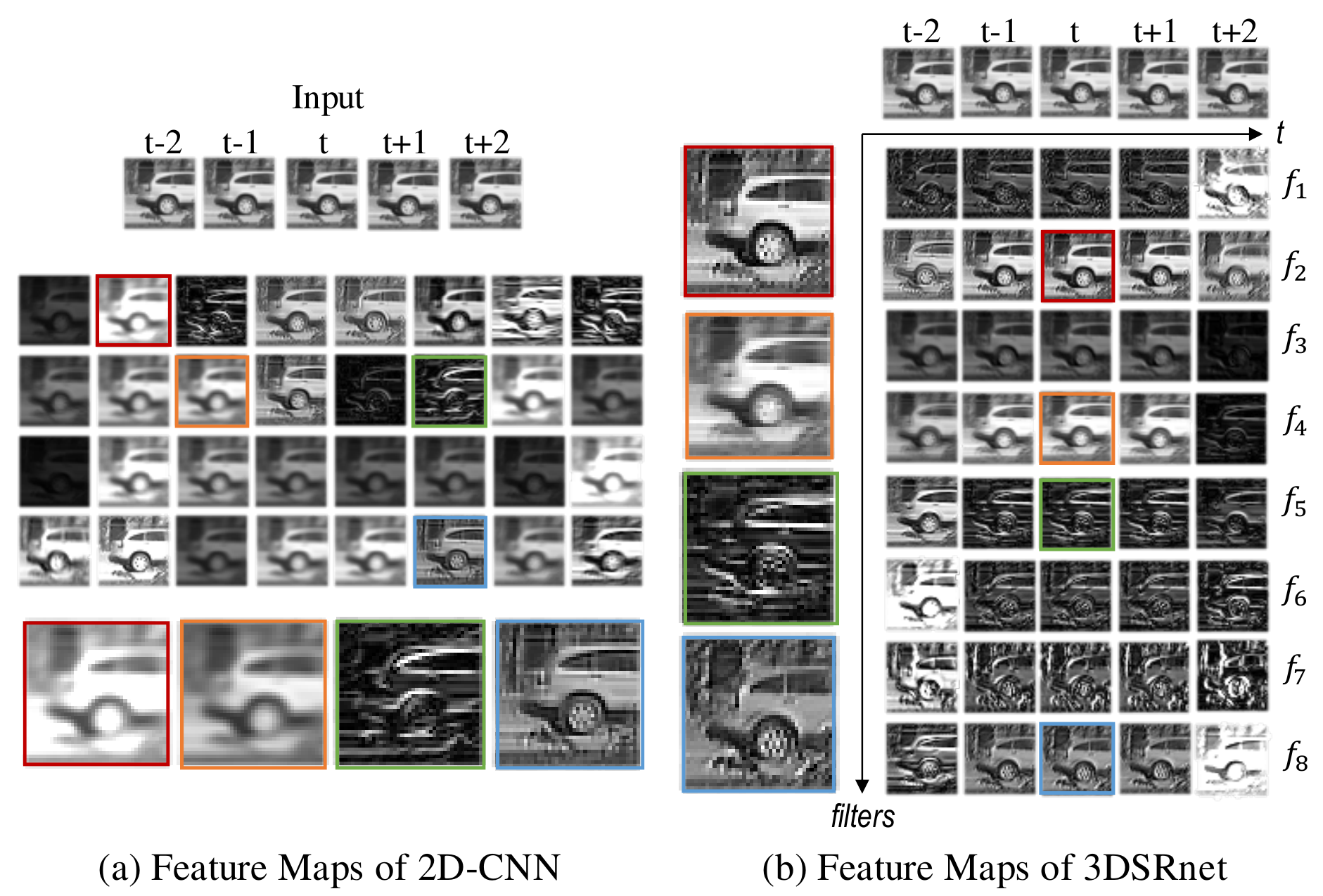}
\caption{Feature map visualization of 2D-CNN and 3DSRnet}
\label{fig:9}
\end{figure}

\subsubsection{Feature map visualizations.} Fig. \ref{fig:9} shows the feature maps produced from the first convolution layer of the 2D-CNN and 3DSRnet, experimented in Table \ref{table:2}. Feature maps of 3DSRnet appears much sharper, due to the shorter time window length of three. 2D-CNN convolves all five frames at the first convolution layer, producing more blurry feature maps. Furthermore, a 3D filter in 3DSRnet produces a GTFM containing five TFMs for each time instant, preserving the temporal information.

\subsubsection{Extrapolation.} The video SR subnet of our 3DSRnet extrapolates the GTFMs with a TFM on their both ends to preserve the temporal depth towards the network's deeper layers. There are different ways of extrapolation such as simply padding them with TFMs filled with zeros or with the outmost TFMs duplicated. However, empirical results showed that there was an insignificant difference in performance with 32.88 dB for duplicate extrapolation and 32.92 dB for zero-filled extrapolation. For simplicity, we choose to use the zero-filled extrapolation.

\begin{table} [b]
\begin{center}
\caption{PSNR (dB) results on each sequence of Vidset4}
\label{table:5}
\scalebox{0.9}{
\begin{tabular} {c|c|c|c|c}
\hline \hline
Vidset 4&\textit{Calendar}&\textit{City}&\textit{Foliage}&\textit{Walk}\\
\hline \hline
$\times$2&27.04&34.13&31.58&36.25\\
$\times$3&24.19&28.25&27.42&30.95\\
$\times$4&22.41&26.81&25.23&28.38\\
\hline \hline
\multicolumn{5}{l}{*trained with the \textit{largeSet}.}\\
\end{tabular}}
\end{center}
\end{table}

\begin{table*}
\begin{center}
\caption{Vidset4 Benchmark-Image and Video SR Methods}
\label{table:4}
\resizebox{\linewidth}{!}{%
\begin{tabular} { c|c|c|c|c|c|c|c|c|c|c|c|c|c|c }
\hline \hline
&\multicolumn{6}{c|}{Image Super-resolution}&\multicolumn{4}{c|}{Video Super-resolution}&\multicolumn{4}{c}{3DSRnet (\textbf{Ours})}\\
\hline
&\multicolumn{2}{c|}{Bicubic}&\multicolumn{2}{c|}{SRCNN \cite{dong2014learning}}&\multicolumn{2}{c|}{VDSR \cite{kim2016accurate}}&\multicolumn{2}{c|}{VSRnet \cite{kappeler2016video}}&\multicolumn{2}{c|}{VESPCN \cite{caballero2017real}}&\multicolumn{2}{c|}{\textit{smallSet}}&\multicolumn{2}{c}{\textit{largeSet}}\\
\hline
$\times$&PSNR&SSIM&PSNR&SSIM&PSNR&SSIM&PSNR&SSIM&PSNR&SSIM&PSNR&SSIM&PSNR&SSIM\\
\hline \hline
2&28.43&0.8685&30.72&0.9176&31.44&0.9257&31.30&0.9278&-&-&31.98&0.9386&\textbf{32.25}&\textbf{0.9410}\\
3&25.29&0.7341&26.54&0.7932&26.84&0.8096&26.79&0.8098&27.25&0.8447&27.64&0.8476&\textbf{27.70}&\textbf{0.8498}\\
4&23.79&0.6342&24.71&0.6923&24.96&0.7121&24.84&0.7049&25.35&0.7557&25.46&0.7498&\textbf{25.71}&\textbf{0.7588}\\
\hline \hline
\end{tabular}}
\end{center}
\end{table*}

\begin{figure*} [h]
\centering
\includegraphics[scale=0.15]{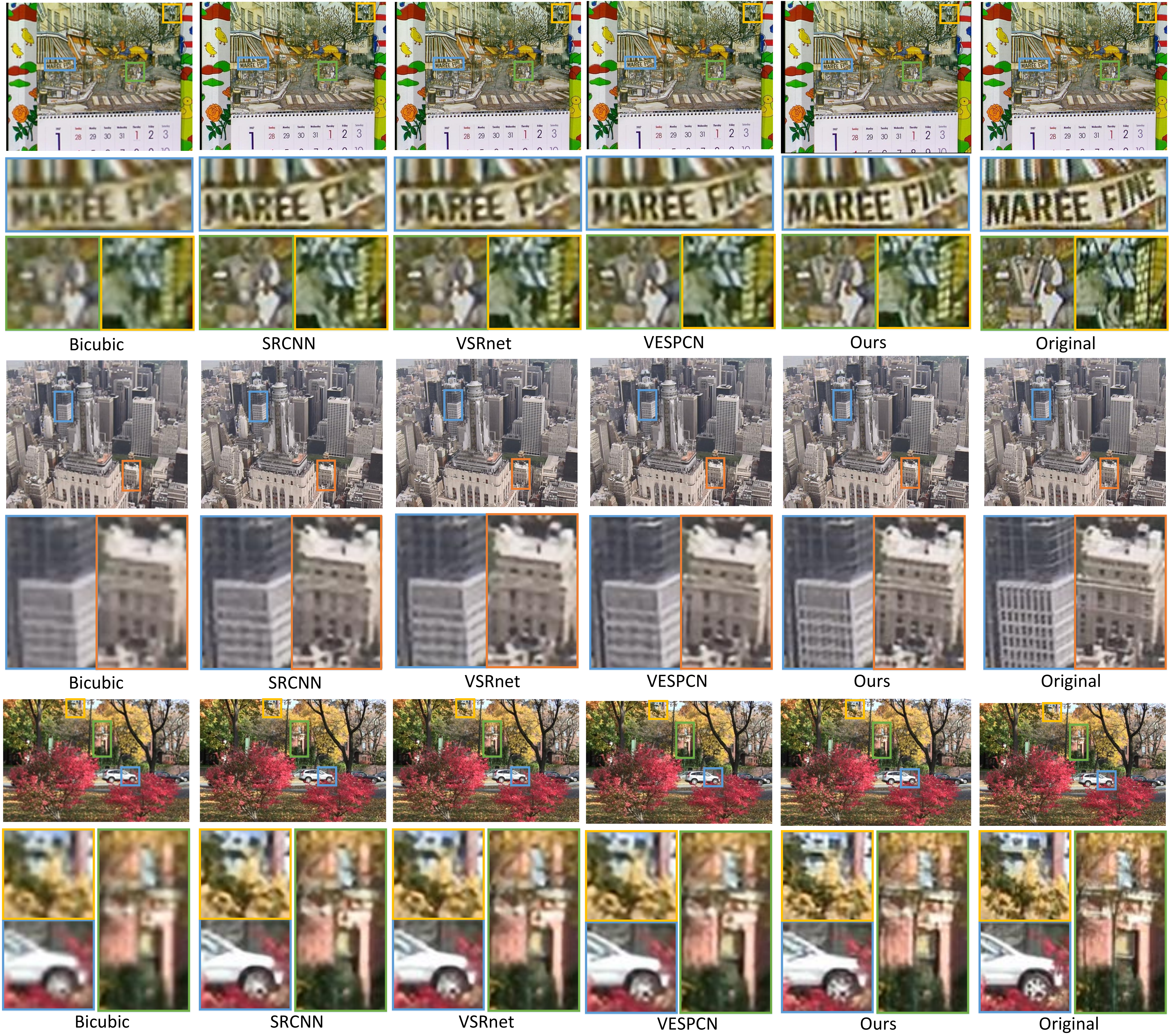}
\caption{Comparisons with the state-of-the-art methods for scale factor 3.}
\label{fig:6}
\end{figure*}

\subsection{Benchmark.} We test the 3DSRnet for quantitative evaluation in comparison with the state-of-the-art image and video SR methods for the Vidset4 dataset - a popular benchmarking test set that contains four video sequences, namely \textit{Calendar}, \textit{City}, \textit{Foliage} and \textit{Walk}. The PSNR comparison against other video SR methods on scale 4 is given in Table \ref{table:3}, and the PSNR and SSIM comparison against image and video SR methods on scale factors 2, 3 and 4 are given in Table \ref{table:4}. The results of video SR methods \cite{caballero2017real,kappeler2016video} are the reported performance on the same test set. The results of \cite{liu2014bayesian}, \cite{liao2015video} and \cite{Liu2017} are from those reported in \cite{Liu2017}. The image SR methods \cite{dong2016image,kim2016accurate} were tested on the set using their respective codes provided by the authors. As shown in Table \ref{table:3} and \ref{table:4}, our 3DSRnet outperforms all the state-of-the-art image and video SR methods. Note that in Table \ref{table:4}, the 3DSRnet shows higher performance with average 0.45 dB and 0.36 dB, respectively for scales 3 and 4 compared to the best performance version (9L-E3-MC) of VESPCN \cite{caballero2017real} which outperformed its 3D-CNN based video SR version. Furthermore from Table \ref{table:5}, our 3DSRnet performs well on all the four sequences without bias toward certain types of videos. Subjective comparisons for the image and video SR methods in Table \ref{table:4} are shown in Fig. \ref{fig:6}.

\begin{figure*}
\centering
\includegraphics[scale=0.35]{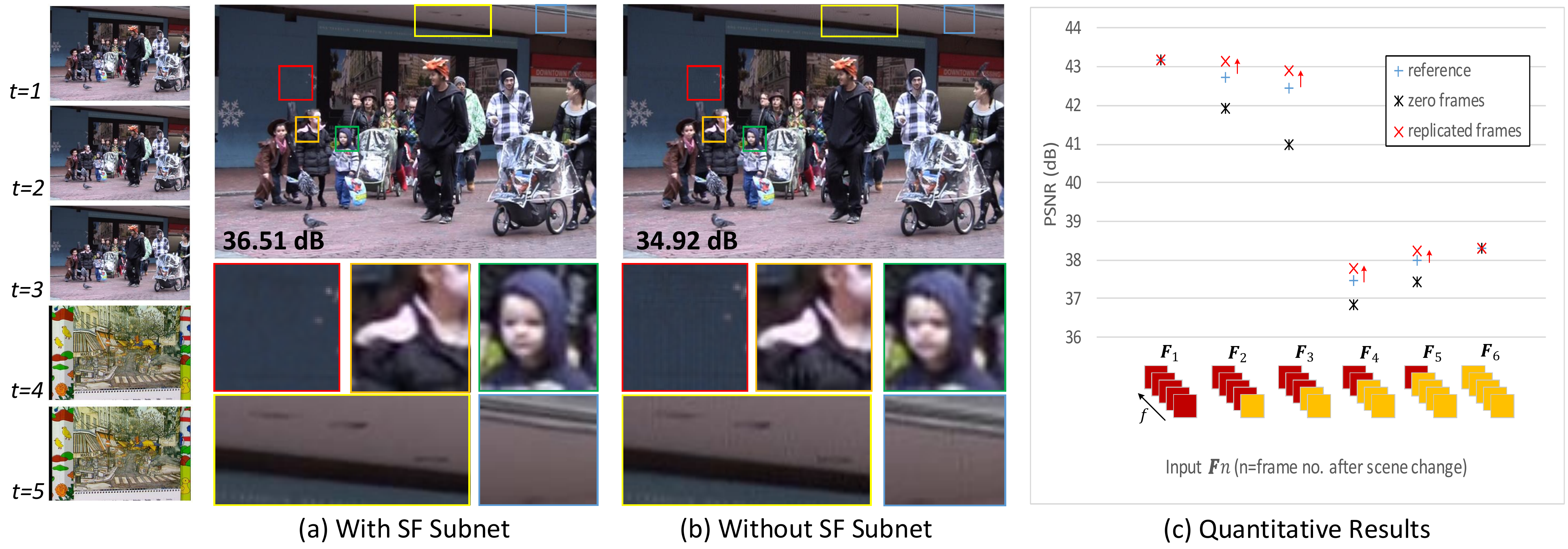}
\caption{Effect of the SF subnet for input with scene change}
\label{fig:8}
\end{figure*}

\subsection{Scale}

Although efficient, a disadvantage of the multi-channel output model \cite{shi2016real} is that separate networks have to be trained for different scale factors since the number of output channels should be $scale^2$ –- scale to the power of 2. Nevertheless, 3DSRnet with a four-channel output for scale 2 can be trained as a single model for different scales. Specifically, for scale 3, the input frames are first up-scaled by 1.5 times using a bicubic filter and then are fed to the 3DSRnet with scale 2. Similarly, for scale 4, the up-scaled input frames of 2 times are used. For training, we use a dataset that contains a mixture of subimages of all scales 2, 3 and 4 denoted as $sub_2, sub_3$ and $sub_4$, respectively, where $sub_3$ and $sub_4$ are up-scaled by 1.5 and 2 times to match the size of $sub_2$. Table \ref{table:6} shows the PSNR performance of the single model trained for all scales 2, 3 and 4 for the Vidset4 dataset. The single model showed the same PSNR performance compared to the separately trained models for scales 2 and 3, but exhibited a slightly higher performance with average 0.2 dB higher in PSNR for scale 4. The single model benefits from the data of various characteristics having diverse frequency ranges even though it is not devoted to learn the training set of a certain scale.

\begin{table} [b]
\begin{center}
\caption{Experiment on scale factors. On the 2nd row are the networks trained separately for each of the scale factors. On the 3rd row are the test results of a single network trained for all scales}
\label{table:6}
\begin{tabular} {c|c|c|c}
\hline \hline
Model&$\times$2&$\times$3&$\times$4\\
\hline \hline
Separate&31.82&27.43&25.27\\
Single&31.82&27.43&25.47\\
\hline \hline
\multicolumn{4}{l}{*PSNR(dB) performance}\\
\end{tabular}
\end{center}
\end{table}

\begin{table} [b]
\begin{center}
\caption{Scene change detection accuracy of 2-layer and 3-layer networks. The test set contains 5,240 samples evenly belonging to the five classes}
\label{table:7}
\begin{tabular} {c|c|c}
\hline \hline
Network&2-Layer&3-Layer \\
\hline
Detection Accuracy (\%)&99.886&99.905\\
\hline \hline
\end{tabular}
\end{center}
\end{table}

\subsection{Scene Change Detection and Frame Replacement Subnet}

Scene change often occurs in video sequences, but little attention has been given in video SR. Without a proper treatment for scene change in input frames, performance degradation is inevitable due to the presence of irrelevant frames. Therefore, in the case of scene change, we swap the unrelated frames with the temporally closest frames of the same scene using the SF subnet introduced in Section 3.2. It improves the quality of the output frames significantly. Fig. \ref{fig:8} shows the qualitative and quantitave results of our 3DSRnet with and without frame replacement for input with scene change. As seen in Fig. \ref{fig:8} (c), if the disparate frames are replaced with zeros, the performance tends to severely drop. 

Let \textit{\textbf{F$_n$}} a series of frames in a sliding time window where the \textit{n}-th frame $f_n$ is the first frame just after scene change. As illustrated Fig. \ref{fig:8} (c), the current frames of \textit{\textbf{F$_1$}}, \textit{\textbf{F$_2$}} and  \textit{\textbf{F$_3$}} correspond to the middle red frames, and those of  \textit{\textbf{F$_4$}}, \textit{\textbf{F$_5$}} and \textit{\textbf{F$_6$}} to the yellow frames. $f_1$ of \textit{\textbf{F$_2$}}, and $f_1$ and  $f_2$ of \textit{\textbf{F$_3$}} are replaced with  $f_2$ of \textit{\textbf{F$_2$}}, and $f_3$ of \textit{\textbf{F$_3$}}, respectively. Similarly, $f_4$ and $f_5$ of \textit{\textbf{F$_4$}}, and $f_5$ of \textit{\textbf{F$_5$}} are replaced with $f_3$ of \textit{\textbf{F$_4$}}, and $f_4$ of \textit{\textbf{F$_5$}}, respectively. Note that \textit{\textbf{F$_1$}} and \textit{\textbf{F$_6$}} do not contain any scene change. As shown in Fig. \ref{fig:8} (c), the PSNR starts to drop from \textit{\textbf{F$_1$}} to \textit{\textbf{F$_3$}}, and from \textit{\textbf{F$_6$}} to \textit{\textbf{F$_4$}}. When the SF subnet is incorporated, the PSNR values without the SF subnet are enhanced by average 0.39, 0.46, 0.32, and 0.25 dB for \textit{\textbf{F$_2$}}, \textit{\textbf{F$_3$}}, \textit{\textbf{F$_4$}} and \textit{\textbf{F$_5$}}, respectively. Table \ref{table:7} shows the detection accuracy of scene change by the SF subnet architecture with two and three layers. Even with the two-layer SF subnet, the detection accuracy of 99.89\% was obtained.

\subsection{Inference Time}

The inference time on an NVIDIA TITAN X GPU is 166 ms and 788 ms for the scale factor of 2 and 4, respectively, to upscale an image of 960$\times$540 resolution from the input of five frames.

\section{Conclusion}

We propose the 3DSRnet, a video SR method that effectively captures spatio-temporal information of LR input frames in reconstructing HR frames throughout deep 3D convolution layers with temporal depth constantly maintained, all without prior motion alignment. The proposed 3DSRnet employs residual learning with the sub-pixel output structure and prevents severe performance drop due to scene change in the multiple input frames by adopting a simple classification network. The experimental results shows that our proposed 3DSRnet outperformed the most state-of-the-art image and video SR methods by maximum 0.45 dB in PSNR.

\end{document}